\documentclass[sigconf]{aamas}

\usepackage{balance} 
\usepackage{amsmath}
\usepackage{graphicx}
\usepackage{subfig}

\setcopyright{ifaamas}
\acmConference[AAMAS '21]{Proc.\@ of the 20th International Conference on Autonomous Agents and Multiagent Systems (AAMAS 2021)}{May 3--7, 2021}{London, UK}{U.~Endriss, A.~Now\'{e}, F.~Dignum, A.~Lomuscio (eds.)}
\copyrightyear{2021}
\acmYear{2021}
\acmDOI{}
\acmPrice{}
\acmISBN{}

\acmSubmissionID{???}

\title[Towards A Multi-agent System for Online Hate Speech Detection]{Towards A Multi-agent System for Online Hate Speech Detection}

\author{Gaurav Sahu}
\affiliation{
  \institution{University of Waterloo}
  \city{Waterloo}
  \country{Canada}}
\email{gaurav.sahu@uwaterloo.ca}

\author{Robin Cohen}
\affiliation{
  \institution{University of Waterloo}
  \city{Waterloo}
  \country{Canada}}
\email{rcohen@uwaterloo.ca}

\author{Olga Vechtomova}
\affiliation{
  \institution{University of Waterloo}
  \city{Waterloo}
  \country{Canada}}
\email{ovechtom@uwaterloo.ca}

\begin{abstract}
This paper envisions a multi-agent system for detecting the presence of hate speech in online social media platforms such as Twitter and Facebook. We introduce a novel framework
employing deep learning techniques to coordinate the channels of textual and image processing.
Our experimental results aim to demonstrate the effectiveness of our methods for classifying online content, training the proposed neural network model to effectively detect hateful instances in the input.
We conclude with a discussion of how our system may
be of use to provide recommendations to users who are managing online social networks, showcasing the immense potential of intelligent multi-agent systems towards delivering social good\footnote{Corresponding author: Gaurav Sahu (gaurav.sahu@uwaterloo.ca)}.
\end{abstract}


\newcommand{\BibTeX}{\rm B\kern-.05em{\sc i\kern-.025em b}\kern-.08em\TeX}

\begin{document}


\pagestyle{fancy}
\fancyhead{}


\maketitle 


\section{Introduction}

A social problem that has become prominent of late is that of
hate speech online.
In this paper, we look at an approach for
detecting instances of these kinds of posts, in contexts such as discussion boards and social networks.
In particular, we examine how designing artificial intelligence algorithms which look at the interplay between visual content and textual content may be especially valuable.
Each channel's processing algorithm can be viewed as an intelligent agent, and determining how best to coordinate the output of these channels into one cohesive interpretation is the multiagent challenge.

In this paper, we sketch an approach to this problem grounded
in the techniques of deep learning.
We contrast this vision with that adopted by other researchers examining how to automate the procedure of hate speech detection.
We also discuss experiments which can demonstrate the advantages of our approach, considering the task at hand as one of effectively classifying the input (either hate speech or not).

From here we move on to reflect on how a multiagent viewpoint
of the task of detecting online hate speech also lends itself
to processing solutions which can be attuned to the user at hand.
In essence, the intelligent agents can incorporate into their
decision making some factors which capture the user's preferences and needs.

Connecting with the theme of this workshop, our research aims to address the uncertainties that arise within social networking environments, where  new content is posted in real time and needs to be assessed and filtered if necessary in a timely manner.
The benefit to society that we strive to achieve is to
enable platform owners to remove harmful content, thereby protecting individuals and preventing online groups from propagating hate speech.


We introduce relevant background for this work in Section~\ref{sec:bg}, formally describe the problem and proposed approach in Section~\ref{sec:method}, and showcase experiments in Section~\ref{sec:experiments}.
Finally, Section~\ref{sec:rest} discusses extended applications of our framework and includes our concluding remarks and scope for future developments in this line of research.



\section{Background}
\label{sec:bg}
Our proposed multi-agent system uses various artificial intelligence techniques and algorithms to extract features from multimodal input and detect hate speech.
In the rest of this section, we discuss relevant background for this work.

\subsection{Deep Learning}
Deep Learning (DL) is a class of machine learning (ML) algorithms that employs multiple layers to extract features from raw input.
Traditional ML algorithms rely on single-layered, simpler architectures such as support vector machines (SVMs)~\cite{cortes1995support} and the probabilistic Naive Bayes model to extract features from given input;
hence, they are unable to capture deep semantic and syntactic information residing in the input.
The last decade has seen a tremendous growth in DL and its applications, and numerous classes of deep neural networks have been proposed to extract features from different types of inputs.
We discuss prominent classes of neural networks proposed to process raw textual and visual input.

\textbf{Recurrent Neural Networks (RNNs).}
A vanilla RNN is a network of \emph{neurons} arranged into successive layers.
Its nodes are connected to form a directed graph along a temporal sequence.
The connections are uni-directional, and every neuron has a real-valued activation function.
Their internal state, better known as \emph{memory}, allows them to process inputs of variable length.
A sentence can be expressed as a variable-length temporal sequence of words, where each word denotes one time-step;
hence, RNNs have been extensively employed for various natural language processing tasks such as machine translation and text classification~\cite{sutskever2014sequence,gillick2015multilingual, jozefowicz2016exploring}.
Long short-term memory (LSTM), a unique RNN that has feedback connections in addition to the standard memory cells is the most widely applied RNN for NLP tasks ~\cite{hochreiter1997long, gers1999learning}.

\textbf{Convolutional Neural Networks (CNNs).}
Convolutional neural networks (CNNs) are most popularly used for visual analysis~\cite{simonyan2014very,szegedy2016rethinking,iandola2016squeezenet,kraus2017automated}.
As the name suggests, CNNs use the convolution operation in at least one of their layers instead of general matrix multiplication in a vanilla multi-layered perceptron (MLP).
While MLPs are prone to over-fitting due to their fully-connected nature, CNNs use small and simple patterns to learn bigger, more complex patterns in data.
This results in fewer connections and lower complexity, making CNNs highly suitable for processing images and videos.
AlexNet~\cite{krizhevsky2012imagenet} and VGG~\cite{simonyan2014very} are two such popular CNNs that gave state-of-the-art performance on the ImageNet classification.
Their deep hierarchical architecture allows them to capture an image's semantic information at various feature-levels.

\textbf{Generative Adversarial Networks (GANs).}
GANs are a specialized class of deep generative models that learn to generate novel data samples from random noise, while matching a given data distribution~\cite{goodfellow2014generative}.
For instance, a GAN trained on a dataset of anime character images can generate novel anime characters, which look highly authentic, at least superficially \cite{chen2019animegan}.
Their flexibility and a wide-range of applications make GANs a popular choice for generation.
Moving past the originally proposed unsupervised learning regime, countless GAN variants successfully adapt it for semi-supervised learning~\cite{salimans2016improved}, fully-supervised learning \cite{isola2017image}, and even reinforcement learning \cite{ho2016generative}.
More advanced applications of GANs include modding in video games~\cite{wang2018esrgan}, motion analysis in video~\cite{vondrick2016generating}, and super-realistic image generation~\cite{stylegan}.
They have even been employed for many text generation tasks such as text style transfer \cite{john-etal-2019-disentangled, Yang_2019_ICCV}.
GANs also demonstrate impressive performance for multimodal tasks such as image captioning~\cite{mirza2014conditional}.

\subsection{Multimodal Deep Learning}
Multimodal Deep Learning (MMDL) involves relating features from multiple modalities (or modes) -- the different sources of information -- such as images, audio, and text.
The goal is to learn a shared representation of the inputs from different modalities, which a neural network may exploit to make intelligent decisions for a desired task.
The earliest attempts to develop such a system involve the work by Ngiam et al. \cite{ngiam2011multimodal}, where sparse Restricted Boltzmann Machines (RBMs) and deep autoencoders were employed to demonstrate the benefit of introducing information from different sources.
Their end-to-end deep graph neural network could reconstruct missing modalities at inference time.
They also demonstrate that better features for one modality can be learned if relevant data from different modalities is available at training time.
However, they also point out that their models could not fully capitalize on the existing information due to heterogeneity in multimodal data.
A line of research dedicated to addressing this issue studies different \emph{fusion} mechanisms--techniques to combine (or \emph{fuse}) information from different modes.
Earlier models such as the bimodal RBMs and Deep Boltzmann Machines (DBMs) use concatenation to fuse cues from different input modes.
While it is a first step towards combining multimodal cues, it results in a shallow architecture~\cite{ngiam2011multimodal,srivastava2012multimodal}.

Zadeh et al.~\cite{zadeh2017tensor} and Liu et al.~\cite{liu2018efficientlow} use Cartesian product and low-rank matrix decomposition instead of concatenation.
Tsai et al.~\cite{tsai2019multimodal} propose multimodal transformers (MulT) which use cross-modal attention mechanisms to combine multimodal cues.
These mitigate the shallowness of the network, while capturing inter- and intra- modal dynamics simultaneously.
However, the resultant architecture either poses a significant computational overhead or further adds to the complexity of a fusion model.
We use GAN-Fusion and Auto-Fusion, two adaptive fusion mechanisms that outperform their massive counterparts on challenging multimodal tasks~\cite{sahu2019dynamic}, described in more detail in Section~\ref{sec:method}.

\subsection{Hate Speech}
Social media platforms have enabled people to connect with others and readily share information.
However, the malicious intent of a few individuals has created a toxic environment online.
One of the prominent social media platforms, Twitter, defines hate speech as follows:

\emph{``Violence against or directly attack or threaten other people on the basis of race, ethnicity, national origin, sexual orientation, gender, gender identity, religious affiliation, age, disability, or serious disease."}

It is, therefore, imperative for such platforms to have an autonomous agent that can flag potential instances of hate speech online.
Such agents can aid in regulating the flow of such content on social media.
Researchers in the AI community have focused on building solutions to address the issue.

\begin{figure*}[ht]
	\centering
	\subfloat[Auto-Fusion]{\includegraphics[scale=0.35,clip,trim=0 -2cm -3cm 0]{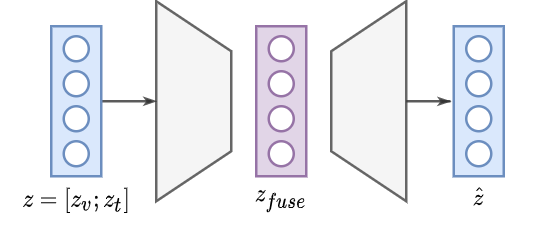}}
	\subfloat[GAN-Fusion module for text]{\includegraphics[scale=0.35,clip,trim=-3cm -2cm 0 0]{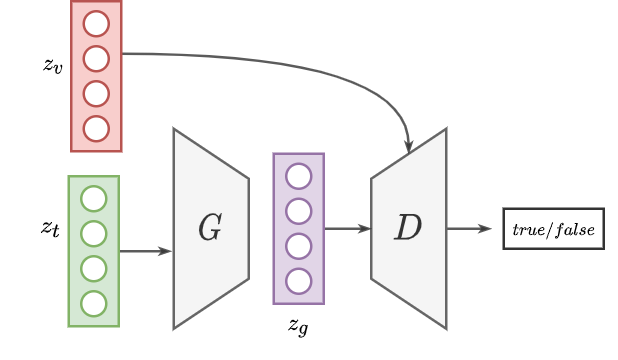}}
	\caption{(a) Auto-Fusion module: First, $\boldsymbol{z}$ (concatenation of $\boldsymbol{z_v}$ and $\boldsymbol{z_t}$) is passed through the autoencoder, which uses $\boldsymbol{z}_{fuse}$, the intermediate latent vector, to output $\boldsymbol{\hat z}$, a reconstruction of input $\boldsymbol{z}$. (b) GAN-Fusion module for text: $\boldsymbol{z_t}$ is passed through a generator (along with a random normal noise), which tries to match its output $\boldsymbol{z_g}$ to $\boldsymbol{z_v}$.
	The discriminator tries to guess the source of its input and outputs a \textit{true/false} label.}
	\label{fig:fusion_models}
\end{figure*}

\textbf{Unimodal setting.}
Schmidt and Wiegand~\cite{schmidt-wiegand-2017-survey} enumerate automated hate speech detection systems in a unimodal setting, where the models leverage textual features such as bag-of-words, word embeddings, and dependency parsing information to detect presence of hate speech online~\cite{chen2012detecting}.
Kwok et al.~\cite{kwok2013locate} and Greevy et al.~\cite{greevy2004classifying} identify the domination of race and sex-based hate speech.
Although surface-level features {such as n-grams} prove helpful, content on social media poses nuanced linguistic challenges like deliberate text obfuscation.
For instance, ``1d10t" is understandable for a human, but it can bypass algorithms relying on keyword spotting to detect hate speech in text~\cite{nobata2016abusive}.
Additionally, the web is full of different but effectively same words.
For instance, ``wow!" and ``wooow!!" carry the same meaning.
Such noisy tokens easily explode the vocabulary size and unnecessarily increase the task's complexity. 
Kumar et al.~\cite{kumar2019online} investigate the use of neural attention mechanisms and gauge the effects of pre-processing to reduce out-of-vocabulary (OOV) instances.

\textbf{Multimodal setting.}
Recently, there have been steady developments towards exploring hate speech detection in a multimodal context.
Kiela et al.~\cite{kiela2020hateful} released The Hateful Meme Challenge, and Gomez et al.~\cite{gomez2019exploring} proposed the MMHS150K dataset.
Both datasets are comprised of an image, its captioned text, and the task is to detect if the image exhibits hate speech or not.
Zhu~\cite{zhu2020enhance}, Muennighoff~\cite{muennighoff2020vilio}, and Velioglu and Rose~\cite{velioglu2020detecting} explore the application of visual-linguistic transformers to extract meaningful cues from images and flag instances of hate speech.
They also confirm the dominance of racist and sexist instances compared to other types of hate speech.
However, these models employ late fusion techniques such as majority voting, which are known to ignore the inter-modal dynamics~\cite{zadeh2017tensor}.

In the next section, we describe our approach, which addresses various issues related to hate speech and multimodal fusion.



\section{Methodology}
\label{sec:method}
We refer to social media posts such as tweets and Facebook posts as \emph{publications}.
Therefore, a publication may consist of an image, text, or a combination of both.

Given an online publication $p$, we denote its visual and textual components by $p_v$ and $p_t$, respectively.
We pose multimodal hate speech detection as a classification problem: given a publication $p$, classify whether it exhibits hate speech or not.
For simplicity, we focus on binary classification in this section, but depending on the experimental settings, the model can very easily be extended for a multi-class classification task.

The following subsections elaborate on different vital components of a multi-agent system to detect hate speech.

\begin{figure}[ht]
  \centering
  \includegraphics[scale=0.4]{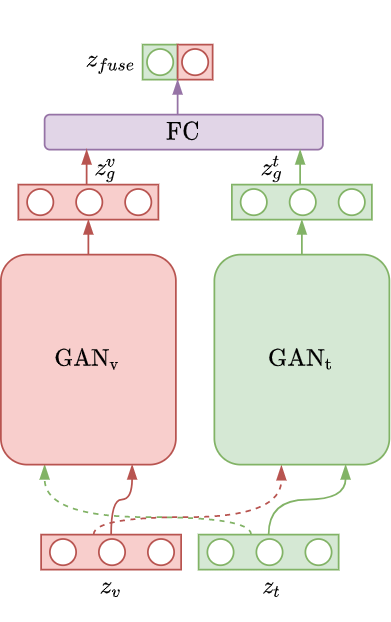}
  \caption{The overall GAN-Fusion architecture. $\textrm{FC}$ denotes the feed-forward layer, which accepts the individual generator outputs and outputs a final fused vector $\boldsymbol{z}_{fuse}$.}
  \label{fig:gan_fusion_compl}
  \Description{Complete GAN-Fusion module}
\end{figure}

\subsection{Visual {(v)} and Textual {(t)} Encoders}
\label{sec:meth_enc}
Given a raw multimodal publication $p$, we first encode $p_v$ and $p_t$ to learn meaningful vectorized representations.
These encoders also serve as the first-level feature extractors.

We use a CNN to encode $p_v$.
More specifically, we use a VGG~\cite{simonyan2014very} module, pre-trained on the ImageNet~\cite{deng2009imagenet} dataset.
Before processing $p_v$, we first transform its dimensionality such that it is a compatible input for the VGG module.
Refer to Section~\ref{sec:experiments} for more experimental details.
We use an RNN to encode $p_t$.
In particular, we use an LSTM cell because it helps in capturing long-range context in a sentence~\cite{hochreiter1997long}.
We also add a layer of word attention mechanism~\cite{luong2015effective} to allow the model to pick up on more important words in a sentence.
The visual and textual encoders output a fixed-dimensional latent vector, denoted by 
${\boldsymbol{z_v}}$ and $\boldsymbol{z_t}$, respectively.
Notably, ${\boldsymbol{z_v}}$ and $\boldsymbol{z_t}$ have an equal number of dimensions.

\subsection{Entity Extraction Modules}
As pointed out in Section~\ref{sec:bg}, content on social media is noisy, which may deteriorate the quality of features extracted by the video and text encoders.
Therefore, we use entity extraction modules to both remove noise from the input and learn a second set of features.

We first clean $p_t$ using the Ekphrasis tool~\cite{baziotis-pelekis-doulkeridis:2017:SemEval2}, which applies advanced text tokenization techniques to process hashtags and emoticons, and also corrects common typographical errors.
For instance, refer to the following example:

``@fiery\_eyes, this is soooo coool borther! {;)} \#coolforever" $\longrightarrow$ ``[user] fiery\_eyes [/user] this is so cool brother! [wink] [hashtag] cool forever [/hashtag]."

Notice that the tokenizer adds appropriate tags for users, hastags and emoticons; corrects spelling mistakes (``borther" $\longrightarrow$ ``brother"); segments concatenated words (``coolforever" $\longrightarrow$ ``cool forever"); and handles elongated texts (``coool" to ``cool").

We also perform part-of-speech tagging (POS-tagging) on the clean text and construct a (subject, object, verb, modifier) tuple for a given sentence.
Additionally, we use Fast R-CNN~\cite{girshick2015fast}, a light-weight object detection module, to identify different acting entities involved in $p_v$.
Extracting entities from both $p_t$ and $p_v$ allows the model to gain a contextual understanding of $p$ as well, and as the model is trained on more examples, it learns to measure the degree of association between a given entity composition (from both image and text) and the presence of hate.

\subsection{Fusion modules}
Since multimodal data is highly heterogeneous, we use two adaptive fusion mechanisms to effectively model inter- and intra-modal dynamics~\cite{sahu2020adaptive, sahu2019dynamic}.
In addition to addressing heterogeneity, these architectures perform impressively on the task of multimodal fusion, despite having significantly fewer number of parameters than the transformers.

\textbf{GAN-Fusion.}
GAN-Fusion{~\cite{sahu2020adaptive}} employs two architecturally similar adversarial modules--one for image and one for text--to fuse latent vectors from different modalities.
Figure~\ref{fig:fusion_models} (b) shows the architecture of $\textrm{GAN}_\textrm{t}$, the GAN-Fusion module for text.
It has two main components: a generator $G$ and a discriminator $D$.
The generator $G$ takes $\boldsymbol{z_t}$ as input along with some random normal noise, and outputs $\boldsymbol{z_g}$, the generated latent vector.
We assign $\boldsymbol{z_g}$ a \emph{false} label and $\boldsymbol{z_v}$ a \emph{true} label.
The discriminator $D$ takes a vector -- either $\boldsymbol{z_v}$ or $\boldsymbol{z_g}$ -- as input.
We denote $D$'s input as $\boldsymbol{z_d}$.

During training, the task of the generator is to match its output latent code $\boldsymbol{z_g}$ as closely as possible to $\boldsymbol{z_v}$.
On the other hand, the discriminator tries to determine if $\boldsymbol{z_d}=\boldsymbol{z_g}$ (in which case, it outputs \emph{false}) or $\boldsymbol{z_d}=\boldsymbol{z_v}$ (in which case, it outputs \emph{true}).
Note that $D$ has no way to know if its input is $\boldsymbol{z_g}/\boldsymbol{z_v}$ beforehand as it only sees a fixed-size vector as input.
To summarize, the generator tries to fool the discriminator, while the discriminator tries to tell apart the difference between its inputs $\boldsymbol{z_g}$ and $\boldsymbol{z_v}$.
Therefore, the overall adversarial objective of $\textrm{GAN}_\textrm{t}$ is given as follows:

\begin{figure*}[ht]
  \centering
  \includegraphics[width=\linewidth]{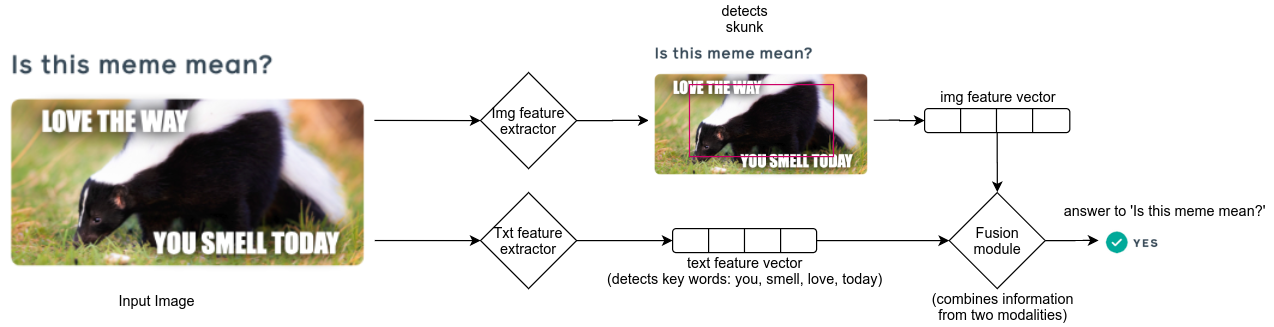}
  \caption{End-to-End pipeline of the proposed multi-agent system (example image from Kiela et al.~\cite{kiela2020hateful})}
  \label{fig:pipeline}
  \Description{End-to-End pipeline of the model}
\end{figure*}

\begin{equation}
\begin{split}
\min_G \max_D J_{adv}^{t} (D, G) &=
\mathbb{E}_{x \sim p_{\boldsymbol{z_{v}}}(x)}[\textrm{log} D(x)] \\
&+ \mathbb{E}_{\boldsymbol{z} \sim p_{\boldsymbol{z_t}} (\boldsymbol{z})}[\textrm{log}(1-D(\boldsymbol{z_g}))]
\end{split}
\label{eq:ganloss_t}
\end{equation}

Equation~\ref{eq:ganloss_t} only describes the objective of $\textrm{GAN}_\textrm{t}$;
however, GAN-Fusion module for the visual modality, $\textrm{GAN}_\textrm{v}$, has the same architecture as $\textrm{GAN}_\textrm{t}$.
It only differs in the input parameters.
Its generator takes $\boldsymbol{z_v}$ as input, and tries to match it with $\boldsymbol{z_t}$.
Therefore, we assign a \emph{true} label to $\boldsymbol{z_t}$ and as before, a \emph{false} label to $\boldsymbol{z_g}$.
The task of the discriminator remains the same: distinguish between the two different types of its inputs ($\boldsymbol{z_t}$ and $\boldsymbol{z_g}$ in this case).
Hence, the objective function for $\textrm{GAN}_\textrm{v}$ can also be expressed in a similar fashion to $\textrm{GAN}_\textrm{t}$ so the overall objective function of the GAN-Fusion module is $J_{adv} = J_{adv}^t + J_{adv}^v$.

Both $\textrm{GAN}_\textrm{t}$ and $\textrm{GAN}_\textrm{v}$ output separate latent codes denoting their respective learned distributions.
Hence, GAN-Fusion uses a feed-forward or a fully-connected layer to combine generated latent vectors of the modality-specific adversarial modules.
It takes the different latent codes as input and outputs a latent vector, $\boldsymbol{z_{fuse}}$, which is finally used for the downstream task of hate speech detection.
Figure~\ref{fig:gan_fusion_compl} shows the complete architecture of the GAN-Fusion module.

\textbf{Auto-Fusion.}
Auto-Fusion{~\cite{sahu2020adaptive}} uses an autoencoder-type architecture to promote information retention during the fusion step. 
It aims to maximize the correlation between the fused and the input latent vectors.
Figure~\ref{fig:fusion_models} (a) shows the architecture of Auto-Fusion module.
It consists of an encoder, which takes the individual modalities' latent codes as input and tries to reconstruct the input using a lower-dimensional latent vector.
The intermediate lower-dimensional latent vector is used as $\boldsymbol{z_{fuse}}$ for the downstream task of hate speech detection.
The reconstruction loss for Auto-Fusion is given as follows:

\begin{equation}
J_{auto} = || \; \boldsymbol{\hat z} - \boldsymbol{z} \; ||^2
\label{eq:autoloss}
\end{equation}

where $\boldsymbol{z}=[\boldsymbol{z_v};\boldsymbol{z_t}]$ is the concatenation of $\boldsymbol{z_v}$ and $\boldsymbol{z_t}$.
In summary, the autoencoder-type framework serves two purposes: 1) compress input to retain only the most important cues from the two modes, and 2) maximize correlation between the fused and the input latent codes.
While Auto-Fusion focuses more on the inter-modal dynamics, it is still a robust architecture for hate speech detection.

\subsection{Final Classification Layer}
After obtaining $\boldsymbol{z_{fuse}}$ from the fusion module, we pass it through a final classification layer.
It is a feed-forward layer, followed by a softmax activation function.
The cross-entropy loss is given by:

\begin{equation}
J_{C} = - \sum_{l \in labels} t(l) \log y(l)
\label{eq:xentropy}
\end{equation}

where $t(\cdot)$ is the ground-truth distribution, $y(\cdot)$ is the predicted probability from the softmax layer.

Figure~\ref{fig:pipeline} shows the end-to-end pipeline of the working model.
First, an input publication from an online social platform, along with its captioned text, passes through the feature extraction step.
Separate extractors for the image and the text modality are used.
The image feature extractor first detects the participating entities involved in the image using object detection techniques.
For instance, in the example shown in Figure~\ref{fig:pipeline}, the image feature extractor first detects the entity `skunk' and outputs a feature vector containing other semantic information about the image.
Similarly, the text feature extractor identifies prominent entities in the text such as the participating subject, object, verb, modifier and the overall sentiment of the text.

In the second step, the text and image features are fused using a fusion module.
This module combines information from the two modalities to cognitively determine if the original input image exhibits online abuse.
The fusion module then outputs a \emph{Hate}/\emph{NoHate} label for binary classification or a valid label for multi-class classification.
Since we train a single end-to-end model for the task, the final objective reflects all the sub-objectives as well:

\begin{equation}
J = J_{C} + J_F
\label{eq:overall_loss}
\end{equation}

where $J_C$ is the classification loss described in Equation~\ref{eq:xentropy}, and $J_F$ is loss of the fusion module.
$J_F=J_{adv}$ or $J_F=J_{auto}$ depending the type of fusion module used.

\section{Experiments and Results}
\label{sec:experiments}
In this section, we discuss different experimental setups to gauge the efficacy of our model for social good.
First, we confirm the effectiveness of our model on the task of speech emotion recognition.
These results serve to confirm the value
of our basic architecture for combining various modalities towards classifying content accurately (and for an ultimate outcome that
also has important social value).
After this, we then show how our model assists with the application in focus, that of hate speech detection.

\textbf{Evaluation Metrics.}
Before describing our classification experiments, we establish the following evaluation metrics:

\begin{itemize}
    \item \textbf{Precision:} It is the ratio of true positives (TP) and the summation of true and false positives (TP+FP).
    For a class, say \textit{Hate}, TP is the set of samples correctly classified as \textit{Hate} and FP is the set of instances where the model incorrectly labels a sample as \textit{Hate}.
    \begin{equation}
        \textrm{Precision}=\frac{\textrm{TP}}{\textrm{TP+FP}}
    \end{equation}
    \item \textbf{Recall:} It is the ratio of true positives (TP) and the summation of true positives and false negatives (TP+FN).
    TP is the same as explained earlier, and FN is the set of instances where the model incorrectly predicts the label to be other than \textit{Hate}.
    \begin{equation}
        \textrm{Recall}=\frac{\textrm{TP}}{\textrm{TP+FN}}
    \end{equation}
    \item \textbf{F1-Score:} It is the harmonic mean of Precision and Recall.
    While Precision indicates the portion of correct predictions, Recall tells us about the total portion of instances actually retrieved.
    F1-score is an important metric as it gauges the \textit{overall} performance of the system.
    \begin{equation}
        \textrm{F1-Score}=\frac{2 \times \textrm{Precision} \times \textrm{Recall}}{\textrm{Precision+Recall}}
    \end{equation}
    \item \textbf{Accuracy:} It is defined as the fraction of true instances out of all the instances.
    \begin{equation}
        \textrm{Accuracy}=\frac{\textrm{TP+TN}}{\textrm{TP+TN+FP+FN}}
    \end{equation}
    Considering the example from before, TN denotes the set of true negatives, i.e., instances where the model correctly labels a sample as not \textit{Hate}.
\end{itemize}
For multi-class classification (>2 classes involved), we compute the above metrics for each class and then use macro-averaging to obtain a final score.
We now discuss our experiments.

\subsection{Speech Emotion Recognition}
\label{sec:emotion}
Speech emotion recognition involves detection of a person's emotional state based on what they say (indicated by the textual modality) and how they speak (indicated by the auditory modality).
It links directly with depression detection, hence, its potential for social good is also immense.
A growing cohort of healthcare professionals use such emotion recognition techniques as an aiding tool to treat patients suffering from clinical depression and PTSD~\cite{he2018automated,rejaibi2019clinical}.
Hence, we perform a comprehensive set of experiments on this task\footnote{{The classes in this case concern emotions, not hate and the channel other than text is audio (not visuals) but the basic multimodal architecture which we promote remains the same.}}.

\begin{table}[ht]
    \small
    \centering
    \begin{tabular}{|c|c|c|c|c|c|c|}
        \hline
         \textbf{Model} & \textbf{Input modes} & \textbf{Fusion type} & \textbf{P} & \textbf{R} & \textbf{F} & \textbf{A} \\
         \hline
         E1 & audio & none & 57.3 & 57.3 & 57.3 & 56.6 \\
         \hline
         E2 & text & none & 71.4 & 63.2 & 67.1 & 64.9 \\
         \hline
         BiL1 & text & none & 53.2 & 40.6 & 43.4 & 43.6 \\
         \hline
         BiL2 & audio+text & Concat & 66.1 & 65.0 & 65.5 & 64.2 \\
         \hline
         E3 & audio+text & Concat & 72.9 & 71.5 & 72.2 & 70.1 \\
         \hline
         MDRE & audio+text & Concat & - & - & - & 71.8 \\
         \hline
         MHA-2 & audio+text & Concat & - & - & - & 76.5 \\
         \hline
         M1 & audio+text & Auto-Fusion & 75.3 & 77.4 & 76.3 & 77.8 \\
         \hline
         M2 & audio+text & GAN-Fusion & \textbf{77.3} & \textbf{79.1} & \textbf{78.2} & \textbf{79.2} \\
         \hline
    \end{tabular}
    \caption{Precision (P), Recall (R), F1-score (F), and Accuracy (A) for emotion recognition on IEMOCAP. {Note: empty cells denote that the metric was not reported in the original paper}}
    \label{tab:emorec_results}
\end{table}

For our experiments, we use the IEMOCAP dataset~\cite{busso2008iemocap}, which originally introduces seven emotion classes, but we only use the \emph{angry, happy, sad} and \emph{neutral} classes, appropriately merging other samples following Sahu~\cite{sahu2020adaptive}.
The dataset provides access to a raw audio vector, its transcribed text, and facial expression of the speaker, but we ignore the facial expressions.

\textbf{Unimodal setting.}
Our unimodal experiments consist of two settings: audio-only and text-only.
For the audio-only setting, we extract eight hand-crafted features from the raw audio input following Sahu~\cite{sahu2019multimodal}.
Then, we use an ensemble of Random Forest (RF), Gradient Boosting (XGB), and multi-layer perceptron (MLP) for classification.
For the text-only setting, we perform two types of experiments.
For the first set of text-only experiments, we use the TF-IDF vectors as features and use an ensemble of RF, XGB, MLP, Multinomial Naive Bayes (MNB), and Logistic Regression (LR) for classification.
For the second set of text-only experiments, we convert raw text into word embeddings and pass them through a bi-directional LSTM with word-level attention (attn) model for emotion detection.

\textbf{Multimodal setting.}
Our multimodal experiments also use two types of approaches.
In the first approach, we concatenate the hand-crafted audio features and the TF-IDF vectors from the unimodal setting, and employ an ensemble of RF, XGB, MLP, MNB, and LR for emotion recognition.

For the second approach, we generate spectrogram images for the raw audio files.
We then use a VGG module for encoding.
For text, we compute word embeddings and use an LSTM encoder identical to our second set of text-only experiments.
Next, we fuse the output signals from the two encoders, followed by a final classification layer.
Note that we do not need an object detection module for the spectrograms.

\begin{table*}[ht]
    \centering
    \begin{tabular}{|c|c|c|c|c|c|c|}
        \hline
         \textbf{Model} & \textbf{Classes} & \textbf{Input modes} & \textbf{Fusion type} & \textbf{P} & \textbf{R} &
         \textbf{F} \\
         \hline
         BiL & binary & text & none & 70.08 & 63.31 & 66.52 \\
         \hline
         TKM & binary & image+text+caption & Concat & - & - & 70.1 \\
         \hline
         SCM & binary & image+text+caption & Concat & - & - & 70.2 \\
        \hline
         FCM & binary & image+text+caption & Concat & - & - & 70.4 \\
        \hline
        \hline
        BiL & multi & text & none & 45.18 & 33.4 & 38.41 \\
        \hline
        BiL & multi & text+caption & none & 45.38 & 33.67 & 38.67 \\
        \hline
        VBiL & multi & image+text+caption & Concat & 55.27 & 35.54 & 43.04 \\
        \hline
        VBiL & multi & image+text+caption & Auto-Fusion & 59.65 & 43.87 & 50.56 \\
        \hline
        VBiL & multi & image+text+caption & GAN-Fusion & \textbf{61.33} &\textbf{ 51.34} & \textbf{55.89} \\
        \hline
    \end{tabular}
    \caption{Precision (P), Recall (R), F1-score (F) for multimodal hate speech detection on MMHS150K. {Note: empty cells denote that the metric was not reported in the original paper}}
    \label{tab:hate_speech_results}
\end{table*}

\textbf{Results.}
We compare the following architectures for speech emotion recognition:
\begin{itemize}
    \item \textbf{MDRE}~\cite{yoon2018multimodal}: The Multimodal Dual Recurrent Encoder baseline employs RNNs to extract features from both audio and text.
    It uses concatenation  to fuse unimodal representations.
    \item \textbf{MHA-2}~\cite{yoon2019speech}: The Multihop Attention Mechanism-2 baseline applies cross-modal attention mechanisms twice to identify the most important tokens for a given audio vector.
    It uses recurrent encoders to obtain latent representations of audio and text, and uses concatenation for fusion.
    \item \textbf{E1}: An ensemble of audio-only RF, XGB and MLP models.
    \item \textbf{E2}: An ensemble of text-only RF, XGB, MLP, MNB, LR.
    \item \textbf{E3}: An ensemble of RF, XGB, MLP, MNB, LR models for the multimodal setting, using concatenation for fusion.
    \item \textbf{BiL1}: The BiLSTM text-only model with attn.
    \item \textbf{BiL2}: A BiLSTM model, which fuses the hand-crafted audio features and TF-IDF vectors using concatenation.
    \item \textbf{M1}: The multimodal system using VGG to encode audio spectrograms and a BiLSTM+attn encoder for text.
    It uses Auto-Fusion as the fusion module.
    \item \textbf{M2}: It has the same architecture as M1 but uses GAN-Fusion to combine features from VGG and BiLSTM+attn.
\end{itemize}
Table~\ref{tab:emorec_results} compiles the performance of the aforementioned models.
We observe that the text-only models (E2, BiL1) are better than the audio-only model (E1).
This may be due to fewer audio features as compared to text.
Expectedly, all models from the multimodal setting outperform the unimodal architectures, except BiL2, which lacks an attn mechanism.
We also note that M2, which uses GAN-Fusion, is the most successful system across all evaluation metrics.
While M2 is the most successful system, both M1 and M2 outperform MDRE and MHA-2, which shows the superiority of our fusion mechanisms over concatenation.

\subsection{Hate Speech Detection}
\label{sec:hate}
Section~\ref{sec:emotion} confirms the effectiveness of our basic architecture towards accurate classification in a multimodal setting.
We now proceed to describe our experiments on hate speech detection.

We use the MMHS150K dataset~\cite{gomez2019exploring}, which contains 150K {labelled} publications, for the hate speech detection experiments.
Each publication features an image, its captioned text, and a textual component;
however, the captioned text may be absent from some samples.
Every sample is assigned to one of the following six classes: \emph{Racist, Sexist, Homophobic, Religion-based, No Hate,} and \emph{Other Hate}.
Despite five categories for hate, the dataset is highly skewed, with $>80\%$ samples belonging to the \emph{No Hate} class\footnote{{Label distribution: No Hate - 81.7\%; Racist - 8.6\%; Sexist - 2.5\%; Homophobic - 2.8\%; Religion-based - 0.1\%; Other Hate - 4.2\%}}.
Hence, we include a two-class variant of the problem in our experiments, where we merge samples from the five hate classes into one \emph{Hate} class.

\textbf{Unimodal setting.}
For unimodal experiments, we only use the textual part of a publication.
We clean the input sentences before computing word embeddings and finally passing them through a BiLSTM+attn encoder.
We also perform POS-tagging on the clean text to obtain a \emph{<subject, object, verb, modifier}> tuple.
The final classification layer exploits this tuple along with the encoder's output $\boldsymbol{z_t}$ to predict a label (e.g. \emph{Hate}, \emph{Not Hate}) for a sample publication.

We first train the model on the relatively easier binary setting, where the model performs a \emph{Hate}/\emph{No Hate} classification.
Keeping the rest of the architecture unchanged, we also run experiments in the full multi-class setting.
To understand the importance of captioned text, we run a separate experiment, where we combine the image's caption (and not the image itself) with the textual component.

\textbf{Multimodal setting.}
In our multimodal experiments, we encode the images using a VGG, which outputs $\boldsymbol{z_v}$.
We then use a Fast R-CNN to detect objects in the image.
Simultaneously, we process the captioned and actual textual component as described in the unimodal setting.
Finally, we combine the visual and textual feature vectors using an appropriate fusion module.
We run experiments on the full multi-class setting, while experimenting with different types of fusion operations.
Section~\ref{sec:method} describes the functioning of various components in more detail and Figure~\ref{fig:pipeline} depicts the final end-to-end pipeline.

\textbf{Results.}
We compare the following systems for hate speech detection:

\begin{itemize}
    \item \textbf{FCM}~\cite{gomez2019exploring}: The Feature Concatenation Model uses an Inception v3 module~\cite{xia2017inception} to encode images and an LSTM layer to encode text.
    As the name suggests, it uses concatenation for fusion.
    \item \textbf{SCM}~\cite{gomez2019exploring}: The Spacial Concatenation Model introduces a new feature map after FCM's Inception module to learn better visual representation.
    \item \textbf{TKM}~\cite{gomez2019exploring}: The Textual Kernels Model trains multiple kernels in addition to the feature maps in SCM to capture multimodal interactions more expressively.
    \item \textbf{BiL}: The unimodal BiLSTM+attn model for encoding text.
    Notably, it uses the Ekphrasis tool for text-cleaning.
    \item \textbf{VBiL}: The multimodal system employs VGG to encode images and a BiLSTM+attn encoder for text.
    It also uses the Ekphrasis tool to clean the textual component.
\end{itemize}

We use Precision, Recall, and F1-Score to evaluate performance of different architectures described above.
Table~\ref{tab:hate_speech_results} compiles the results of our hate speech experiments~\footnote{{Due to the dataset being more highly skewed, accuracy is not listed.}}.
It includes information about input modalities involved, the experiment setting (binary/multi) and the type of fusion module used.
These results are preliminary, and form part of our work in
progress on this solution.
For binary classification experiments, we first observe that unimodal BiL is highly competitive with the multimodal FCM, SCM, and TKM baselines.
Since LSTM is common to all of them, we attribute the impressive performance to text cleaning and the powerful attention mechanism.
The multi-class experiments show an expected drop in performance.
This is also partly due to a high class imbalance in the dataset.
Interestingly, image captions have no significant effect by themselves on the performance of the model (as shown by BiL's multi-class experiments).
However, inclusion of images (rows where `Inputs modes' is image+text+caption) boosts classification performance.
This shows that our effort to reason about both text and image
together in order to detect hate speech has been effective~\footnote{{These results are preliminary. Next steps include running VBiL in a binary setting and using more powerful encoders for text and images}}.

\section{Conclusion, Discussion and Future Work}
\label{sec:rest}

As presented in Figure~\ref{fig:pipeline}, the two primary modalities of text and image processing need to coordinate their activities to effectively detect possible hate speech in online networks.
The architecture that we have presented here offers a novel direction for integrating these two agents for classifying content as hate.
This multiagent system will serve to inform users invested in adjusting the presentation of content online (be it to flag what might be questionable or to remove it from the stream) and as such this work is therefore focused on the overall aim of improving social good (mitigating harm inflicted by users who are not pro-social).

\begin{figure*}[ht]
  \centering
  \includegraphics[width=0.75\linewidth]{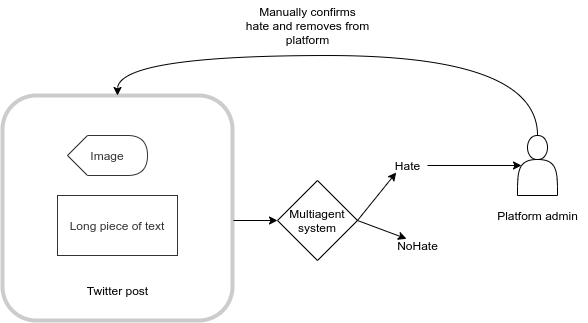}
  \caption{How to use the proposed multiagent system to moderate content online.}
  \label{fig:user_agent}
  \Description{Envisioned usecase}
\end{figure*}

Figure~\ref{fig:user_agent} shows our envisaged design of the process to moderate content.
The platform admin is the end user, with all required privileges
to control the inflow and outflow of content on the platform.
She can add new content to the platform or
edit/delete existing material at her discretion.
Once our system detects potentially hateful content, the platform admin can be notified.

We view our connection to agent-based processing to be aligned with
those of other researchers who seek to yield improvements in the social conditions of individuals, once more effective processing of the online environments of these users has been achieved.
This approach has been adopted, for instance, in work published
at AAMAS regarding delivering better influence maximization in uncertain social networks in order to assist homeless youths (e.g. Kamarthi et al.~\cite{kamarthi2019influence}, Yadav et al.~\cite{yadav2016using}) where the algorithms designed, once run, can help to propose recommendations for actions in the real-world which will achieve social good.

To summarize, in this paper, we sketch a solution to the problem of hate speech detection through a multi-agent system.
Section~\ref{sec:experiments} demonstrates the effectiveness of the architecture proposed for coordinating text processing
and image processing and offers an improved analysis of
online content.
We show our model works for a key classification task (that also has social value):
effectively detecting emotion.
The central value of using a combination of modalities in this application area is the kind of promise that we seek to build upon in order to deliver
new advances for the very important social problem of online hate
speech detection.
The results of Section~\ref{sec:hate} present the
current analysis of our algorithms within this particular context.
We see avenues for future work from this particular work in progress,
in order to refine the processing towards greater performance.

\balance

\textbf{Improved GANs.}
Despite the high-quality generation, GANs suffer from the ``mode-collapse" problem, wherein they fail to capture entire modes in the real data distribution.
For instance, a GAN trained on the MNIST dataset -- a collection of handwritten digits from one to ten -- might neglect a subset of digits from its output.
Furthermore, training GANs can be tricky sometimes~\cite{arora2017generalization,sinn2018non}.
An interesting line of solutions includes an implicit maximum likelihood estimation training objective~\cite{Li2018OnTI, Li2018ImplicitML, pacgan}, which insists on the use of full-recall GANs.

\textbf{Embedding External Knowledge of the World.}
Modelling context is crucial for multimodal hate speech detection.
In many cases, neither the text nor the image are hateful individually, but acquire such meaning in combination. These challenging instances require the system to possess an external knowledge of the world.
Therefore, dynamic knowledge sources such as Google Vision's Web Entity Detection API\footnote{\url{https://cloud.google.com/vision/docs/detecting-web}}, which can tackle the ever growing pool of information on the web, can immensely enhance the performance of the system.
Another line of approaches includes effectively exploiting the stored knowledge in the neural network's parameters~\cite{lewis2020retrieval}.







\bibliographystyle{ACM-Reference-Format} 
\bibliography{sample}


\end{document}